\documentclass[]{article}
\usepackage{proceed2e}

\usepackage{graphicx} 
\usepackage{subfigure} 

\usepackage{natbib}

\usepackage{amsmath}
\usepackage{bm}
\usepackage{comment}

\renewcommand{\v}[1]{\mbox{\boldmath{${#1}$}}}
\newcommand{\qq}[1]{``{#1}''}

\newcommand{\tablespace}{\rule[0in]{0in}{0.20in}}

\title{Learning Densities Conditional on Many Interacting Features}

\author{ {\bf David C. Kessler} \\  
Department of Biostatistics \\
University of North Carolina \\
at Chapel Hill
\And 
{\bf Jack Taylor}  \\ 
National Institute of \\
Environmental Health Sciences   \\ 
\And 
{\bf David B. Dunson}  \\ 
Departments of Statistical Science \\
and Electrical \\
and Computer Engineering          \\ 
Duke University \\              
} 

\begin{document} 
 
\maketitle 
\begin{abstract} 
Learning a distribution conditional on a set of discrete-valued features is a commonly encountered task.  This becomes more challenging with a high-dimensional feature set when there is the possibility of interaction between the features.  In addition, many frequently applied techniques consider only prediction of the mean, but the complete conditional density is needed to answer more complex questions.  We demonstrate a novel nonparametric Bayes method based upon a tensor factorization of feature-dependent weights for Gaussian kernels.  The method makes use of multistage feature selection for dimension reduction.  The resulting conditional density morphs flexibly with the selected features.
\end{abstract} 

\section{MOTIVATION}
\label{S:Motivation}
\begin{figure}[ht]
\vskip 0.2in
\begin{center}
\centerline{\includegraphics*[width=\columnwidth,viewport=0 130  600 650]{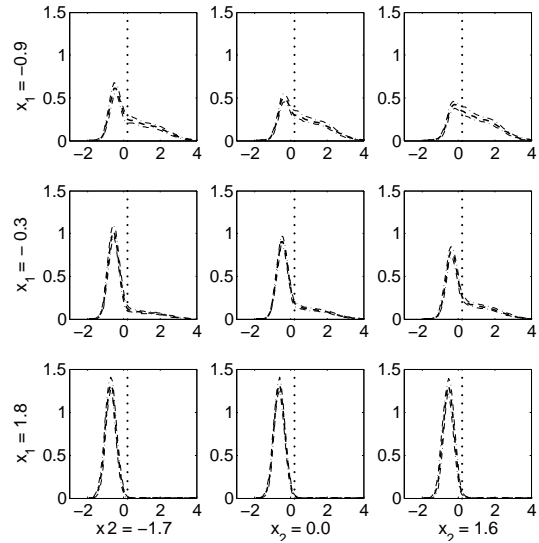}}
\vskip 0.1in
\caption{Conditional Densities With Similar Means But Feature-dependent Higher Moments; \citep{chung_dunson_2009}.}
\label{F:wacky}
\end{center}
\vskip -0.2in
\end{figure}

\begin{figure}[ht]
\vskip 0.2in
\begin{center}
\centerline{\includegraphics*[width=\columnwidth,viewport=40 40 500 450]{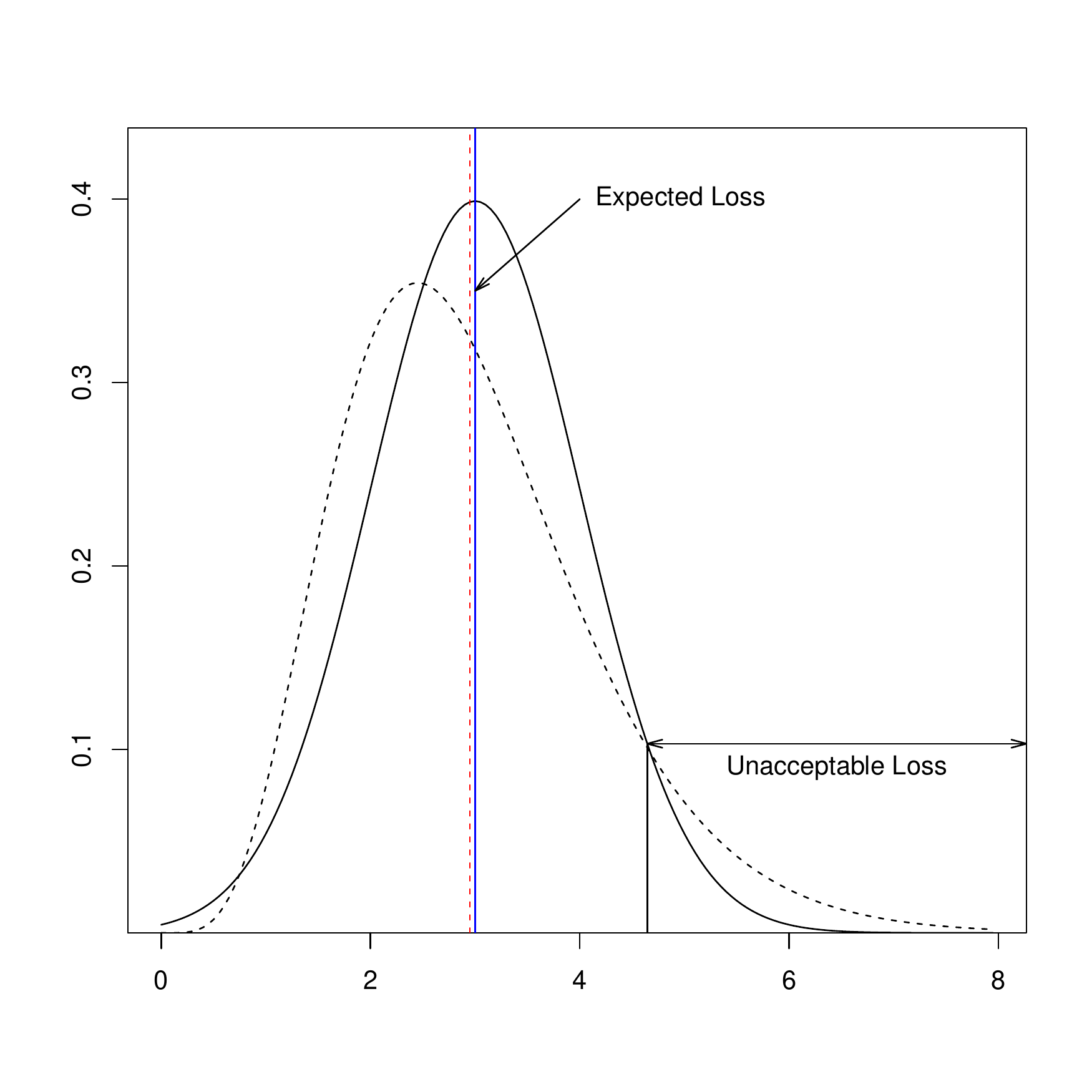}}
\vskip 0.1in
\caption{Conditional Distributions With the Same Mean But Different Tail Probabilities.}
\label{F:tailprob}	
\end{center}
\vskip -0.2in
\end{figure}

Many areas of research are concerned with learning the distribution of a response conditional on numerous categorical (discrete) features.  The features that have actual importance for the characterization of this distribution are not usually known in advance, and in many cases hundreds or thousands of features are associated with each response.  In addition, it is frequently the case that the features interact in complex ways.  Methods that attempt to consider each potential interaction can quickly become mired in the enormous space of models.  For example, in a moderate-dimensional case involving $p=40$ categorical features, each with $d_j=4$ possible realizations, considering all possible levels of interaction leads to a space of $4^{40} \approx 10^{24}$ possible models.  Parallelization and technical tricks may work for smaller examples, but data sparsity and the sheer volume of models force us to consider different approaches.
In addition, approaches to learning conditional densities that are based on mean regression do not always consider the variation in form of the density.
That is, the conditional density may vary in more than just location, as illustrated in Figure \ref{F:wacky}.
Methods that score well for measures based upon mean square prediction error (MSPE) may fall short on other important questions.  As shown in Figure \ref{F:tailprob}, it may be the case for two distinct combinations $\v{x}^{(1)}$ and $\v{x}^{(2)}$ of features that $E(y | \v{x}^{(1)}) = E(y | \v{x}^{(2)})$ but that $P(y>c|\v{x}^{(1)}) \gg P(y>c|\v{x}^{(2)})$.  Such differences in the predicted probability of extreme observations can be of considerable interest in environmental, financial, and health outcomes settings.
In the work that follows, we present a novel nonparametric Bayes (NPB) approach to learning conditional densities that makes use of a conditional tensor factorization to characterize the conditional distribution given the feature set, allowing for complex interactions between the features.  The proposed method incorporates distinct variable selection steps to address the challenges of high-dimensional data and produces conditional density estimates that allow assessment of tail risks and other complex quantities.

\section{BACKGROUND}
\vskip -6pt
The primary goal for our work is to model
the conditional density $f(y|\v{x})$, where the form of this density for the
response $y$ changes flexibly with the feature vector $\v{x}$.
There is a large body of work devoted to this idea of density regression in settings
involving $\v{x}$ of dimension $p \le 30$, and such models have provided many
options for that situation.  We wish to develop techniques for problems
involving much larger $p$, and ideally to scenarios where $p > 1,000$.  While
several techniques exist for this high-dimensional setting, they can result in black-box models
that do not motivate understanding of the effect of a particular feature on the
response.  We want to provide a method that performs variable selection,
assesses the probability of a feature's inclusion in the model, and provides
easily interpretable estimates of the impact of different features.

This
classically nonparametric problem has been addressed with variations
on the finite mixture model, summarized in its general form here:
\begin{align}\label{E:KernMix}
    f(y) = \sum_{h=1}^K \pi_h \mathcal{K}(y;\:\theta_h).
\end{align}
This is the basic form of the hierarchical mixture of experts model (HME, \citet{jordan_jacobs_1994}). In this representation, $K$ represents the number of contributing parametric
kernels $\mathcal{K}(;\theta_h)$ distinguished by parameters $\theta_h$.  The
$\pi_h$ provide the weights in this convex combination of kernels, where
$\sum_{h=1}^K \pi_h = 1$ and $(\pi_1,\ldots,\pi_K) \in \mathcal{S}_{K-1}$, the $K-1$
probability simplex.  The most straightforward forms rely on a prespecified $K$
and include the features $\v{x}$ in a linear model for the mean.  HME methods in the frequentist literature have often relied on expectation maximization (EM) \citep{dempster_etal_1977} techniques, which can suffer from overfitting \citep{bishop_svensen_2003}.  EM approaches in the Bayesian literature seek to avoid this; \citet{waterhouse_etal_1996} employed EM to find maximum a posteriori (MAP) estimates, using the inherent Bayesian penalty against complexity to regulate those estimates.  In addition, the Bayesian framework allows the quantification of uncertainty about the parameters in the model.

The advent of
nonparametric Bayes (NPB) techniques like the Dirichlet process (DP) prior
prompted techniques like that in \citet{muller_etal_1996},
which focused on flexible conditional mean regression through joint modeling of the response and features.  Several
subsequent methods have used the features to inform the weights $\pi_h$, and
implemented this using dependent Dirichlet Process (DDP) mixtures.
\Citet{deiorio_etal_2004} proposed an ANOVA DDP model with the same weights
that used a small number of multiple categorical features to index random
distributions for the response; in this development the weights $\{\pi_h\}$
were assumed to be the same across the contributing mixtures.
\citet{griffin_steel_2006} developed an ordered DDP, where the feature vectors
were mapped to specific permutations of the weights $\{\pi_h\}$, yielding
different density estimates for different feature vectors.
\citet{reich_fuentes_2007} and \citet{dunson_park_2008} employed the kernel
stick-breaking process to allow features to influence the weights.
\citet{chung_dunson_2009} presented a further alternative in the probit
stick-breaking process, which uses a probit transform of a real-valued function
of the features to incorporate them into the weights.  Methods that use joint modeling of response and features \citep{shahbaba_neal_2009,hannah_etal_2011,dunson_xing_2009} are popular and can work well under many circumstances, but in other settings the estimation of the  marginal distribution of the features becomes a burden.

While the discrete
mixture approach (both finite and infinite) has provided the bulk of techniques
for Bayesian density regression, there are notable exceptions.  For example,
\citet{tokdar_etal_2010} developed a technique based upon logistic Gaussian
processes.  \citet{jara_hanson_2011} presented an approach using mixtures of
transformed Gaussian processes.

These and other methods of Bayesian density regression have proven successful,
but as data sets have grown in size
and complexity, these approaches encounter difficulties.  One particular
challenge derives from the so-called \qq{curse of dimensionality} - that is, as
we consider problems in higher and higher dimensions, where we consider larger
and larger feature vectors, the complexity of interaction between
these explanatory variables grows explosively and data sets may only sparsely fill
the associated space.  This is even more daunting when we consider discretely valued features, since we must consider the factorial combinations of those levels.

The associated challenges of variable selection and dimensionality reduction
have been explored in Bayesian density regression.  
Dimensionality reduction has
a goal similar to that of variable selection, that of finding a minimal set of
features that account for variation in the response.  The logistic Gaussian
process approach of \citet{tokdar_etal_2010} includes a subspace projection
method to reduce the dimension of the feature space.  \citet{reich_etal_2011}
developed a technique for Bayesian sufficient dimensionality reduction based
upon a prior for a central subspace.  While all of these approaches have
demonstrated their utility, they do not scale easily beyond $p=30$ features.

There are also techniques like the random forest \citep{breiman_2001} that aim to
find parsimonious models for density estimation involving a large number of
features. One disadvantage to this type of \qq{black box} method is in interpreting the impact of specific features on the response.  Bayesian additive regression trees (BART) \citep{chipman_etal_2006,chipman_etal_2010} focus on modeling the conditional mean and assume a common residual distribution.  As previously noted, there are many questions that require learning about more than just the conditional mean of the response.

To address these disparate challenges, we propose an approach based upon a 
conditional tensor factorization (CTF) for the mixing weights. As
in the DDP and certain of the kernel stick-breaking methods, the features
influence the mixing weights for this CTF model.  The conditional tensor factorization facilitates borrowing of information across different profiles in a flexible representation of the unknown density.  We focus our attention on
situations involving continuous responses and categorical features.
\section{MODEL}
\label{S:Model}
We consider a univariate response $y$ and a vector of categorical features $x = (x_{1}, \ldots, x_{p})$, where the $j^{th}$ feature $x_{j}$ can take on the values $1, \ldots, d_j$.

We would like a model that can flexibly accommodate conditional densities that change in complex ways with changes in the feature vector.  In addition, we must 
consider situations where the number of possible $\v{x}$ approaches or exceeds the number of observations.  In this setting, there may be very few or no exemplars for certain feature vectors.  This sparsity can derail methods that rely on the complete feature vector $\v{x}$ for learning about the conditional distribution of the response.  To address this, we propose a Tucker factorization style of approach and the following general model for the conditional density $f(y|x)$:
\begin{align}
f(y_i | \v{x}_i) &= \sum_{h_1=1}^{k_1} \cdots \sum_{h_p=1}^{k_p} \pi_{h_1,\!\cdots,\!h_p}(\v{x}_i) \: \lambda(y_i;\theta_{h_1,\!\cdots,\!h_p}) \notag
\end{align} 
where
\begin{align}
\label{E:SoftMap}
\pi_{h_1,\cdots,h_p}(\v{x}) &= \prod_{j=1}^p \pi^{(j)}_{h_j}(x_j)
\end{align}
Each $\pi^{(j)}$ can be visualized as a matrix, where the row indexed by $x_j$ contains weights for the combinations of observed feature $x_j$ and latent features $h_j=1,\ldots,k_j$.  The weights for a particular value of $x_j$ are constrained to be $\in [0,1]$, and $\sum_{h_j=1}^{k_j} \pi^{(j)}_{h_j}(x_j) = 1$.
This is similar in spirit to the classification approach proposed by \citet{yang_dunson_ctf_2012}, but the method presented here focuses on regression and density estimation problems.  Tucker decompositions \citep{tucker_3fact_1966} and other kinds of decompositions have appeared in the machine learning literature before. \citet{xu_etal_2012} developed an \qq{infinite} Tucker decomposition, making use of latent Gaussian processes rather than explicit treatment of tensors and matrices; in comparison, the proposed method uses the Tucker decomposition to characterize the mapping of features into weights.  Other factorizations have been used for similar problems; \citet{hoff_multiway_2011} presented a reduced-rank approach for table data, but this approach focused on the development of estimates for the mean of a continuous response.  \citet{chu_ghahramani_2009} derives an approach for partially observed multiway data based upon a Tucker decomposition; in this case the objective is to learn about the latent factors driving observations rather than the characterization of the response distribution.

The collection across $j=1,\ldots,p$ forms a \qq{soft} clustering from the
$d_1 \times \cdots \times d_p$ dimensional space of the observed $\v{x}$ to a
potentially smaller $k_1 \times \cdots \times k_p$-dimensional space.  That is, a feature vector $\v{x}$ is not exclusively associated with a single kernel, but  rather with all $k_1 \times \cdots \times k_p$ kernels through the corresponding weights.

This form for the mixing weights allows borrowing of information across different combinations of $h_1,\ldots,h_p$.  Learning about the density conditional on a sparsely observed feature vector $\v{x}^{(*)}$ does not rely exclusively on observations with that feature vector; instead, each observation contributes some information.  The impact of non-matching feature vectors is governed by the set of maps $\pi^{(j)}$, rather than some hard classification.  In settings of extreme sparsity, where most feature vectors are not represented, this is an attractive property.
This uses many fewer parameters than a full factorial representation, and is still flexible enough to represent complex conditional distributions.

Finally, we assume normal kernels for the $\lambda$, yielding:
\vskip -0.3in
\begin{align}\label{E:CTFModel}
\: & f(y_i|\v{x}_i) \notag \\
\: & = \sum_{h_1=1}^{k_1} \! \cdots \! \sum_{h_p=1}^{k_p} \Big\{ N(y_i;\theta_{h_1,\cdots,h_p}, \tau_{h_1,\cdots,h_p}^{-1}) \notag \\
\: & \hskip 1.5in \times \prod_{j=1}^p \pi^{(j)}_{h_j}(x_{ij})\Big\} 
\end{align}
\vskip -0.1in
This resembles other mixture-based approaches to density estimation as
originally specified in \eqref{E:KernMix}, but the proposed model for the weights provides the desired support for sparsity and information borrowing previously discussed.

\section{METHODS}
\vskip -0.1in
We consider two primary tasks in learning the conditional distribution.  The first is to identify those features which provide the most information about the response, and the second is to learn the form of the conditional distribution given the set of informative features.  Both tasks will be influenced by our prior assumptions about uncertainty in the model parameters, quantified as prior distributions.  For computational convenience, we employ conjugate priors where possible.

The model proposed in \eqref{E:CTFModel} can be augmented to give a complete-data likelihood assuming a specific classification vector $z_i$ for each observation, kernel mean parameters $\theta_{h_1,\cdots,h_p}$, kernel precision parameters $\tau_{h_1,\cdots,h_p}$ and the soft-clustering parameters $\pi^{(j)}$:
\begin{align}
\label{E:AugLike}
\prod_{i=1}^N 
    \prod_{h_1=1}^{k_1} \cdots \prod_{h_p=1}^{k_p}     
    \Big\{
    N(y_i; \theta_{h_1,\cdots,h_p}, \tau_{h_1,\cdots,h_p}^{-1}) \times \notag \\
	\prod_{j=1}^p \pi^{(j)}_{h_j}(x_{ij})
    \Big\}
    ^{1[z_i=(h_1,\cdots,h_p)]}
\end{align}
The dimension of the full vectors $\v{\theta}$ and $\v{\tau}$ will be denoted by $M$, where $M=k_1 \times k_p$.

\subsection{PRIOR STRUCTURE}
\begin{enumerate}
\item $\theta_{h_1,\cdots,h_p} \sim N(0, \tau_0^{-1})$.
\item $\tau_{h_1,\cdots,h_p} \sim \text{Gamma}(\delta_t/2, \gamma_t/2)$
\item $\pi^{(j)}(x_j) = (\pi^{(j)}_1(x_j), \ldots, \pi^{(j)}_{k_j}(x_j)) \sim$ \\
$\text{Dir}(\frac{1}{k_j}, \ldots, \frac{1}{k_j})$ \\
for $j=1,\ldots,p \text{ and } x_j = 1, \ldots, d_j$
\item $\tau_0 \sim \text{Gamma}(\delta_0/2, \gamma_0/2)$
\end{enumerate}

The final set of parameters, the $k_1, \ldots, k_p$, present a particular challenge.  Since each $k_j$ can take on the values $1, \ldots, d_j$, the resulting discrete space can be immense, and including these as parameters in the sampler is not an attractive option.  Instead, we develop a stochastic search variable selection (SSVS) step that makes use of a \qq{hard} clustering to evaluate different $k_j$ values.

\subsection{FULL CONDITIONALS}
\label{S:FullCond}
Given the augmented likelihood in \eqref{E:AugLike}, the assumed prior distributions and fixed values $k_1, \ldots, k_p$, the full conditional distributions are:
\begin{enumerate}
\item $\theta_{h_1,\cdots,h_p}|\cdots \sim N(\mu_{h_1,\cdots,h_p}^{*}, (\tau_{h_1,\cdots,h_p}^{*})^{-1})$, where: \\
      $\tau_{h_1,\cdots,h_p}^{*} = \tau_0 + \tau_{h_1,\cdots,h_p} \sum_{i=1}^N 1[z_i=(h_1,\cdots,h_p)]$ \\
      $\mu_{h_1,\cdots,h_p}^{*} = $ \\
      $\{\tau_{h_1,\cdots,h_p} \sum_{i=1}^N y_i 1[z_i=(h_1,\cdots,h_p)]\} / \tau_{h_1,\cdots,h_p}^{*}$
\item $\tau_{h_1,\cdots,h_p}|\cdots \sim \text{Gamma}(\delta^{*}/2,\gamma^{*}/2)$, where:
$\delta^{*} = \delta_t + \sum_{i=1}^N 1[z_i=(h_1,\cdots,h_p)]$ \\
$\gamma^{*} = \gamma_t + \sum_{i=1}^N 1[z_i=(h_1,\cdots,h_p)] (y_i - \theta_{h_1,\cdots,h_p})^2 $ \\
\item $\tau_0 | \cdots \sim \text{Gamma}([\delta_0+M]/2, [\gamma_0+\v{\theta}^T \v{\theta}]/2)$
\item $(\:\pi^{(j)}_{1}[x_j], \ldots, \pi^{(j)}_{k_j}[x_j]\:) | \cdots $
\begin{align*}
\sim \text{Diri}( & 1/k_j + \sum_{i=1}^N 1[z_{ij} = 1   ], \: \ldots, \\
               \: & 1/k_j + \sum_{i=1}^N 1[z_{ij} = k_j] )
\end{align*}
\item 
$\text{Pr}[z_i = z_{jm}^* \equiv (h_1,\ldots,h_{j-1},m,h_{j+1},\ldots,h_p)] | \propto $ \\
\[
\phi\Big[(y_i - \theta_{z_{jm}^*})\sqrt{\tau_{z_{jm}^*}}\Big] \times
  \pi^{(j)}_{m}(x_{ij})
\]
for $m=1,\ldots,k_j$ within each $j=1,\ldots,p$.
\end{enumerate}
The updates for $\v{\theta}$, $\v{\tau}$ and $\v{\pi^{(j)}}$ can be done blockwise.  The $z_i$ can updated blockwise at each position $j$.

\subsection{FEATURE SELECTION}
To learn appropriate values for $k_1, \ldots, k_p$, we use a feature selection step based upon a special form of the $\pi^{(j)}$.  This special form of the mapping in \eqref{E:SoftMap}
results if exactly one of the elements of $\pi^{(j)}(x_j)$ is equal to 1, with the other $k_j-1$ elements equal to zero.  This gives a \qq{hard} clustering of each feature vectors $\v{x}_i$ to exactly one element of the $M-$dimensional space outlined above.  Given a particular clustering and the prior structure outlined above, we can approximate a marginal likelihood for that clustering; these marginal likelihoods provide calibrated measures of different clusterings that drive a stochastic search.
We make the simplifying assumption that $\tau_0 = \tau$ and retain the $\text{Gamma}(\delta_t/2, \gamma_t/2)$ prior for $\tau$.  This gives an exact form for the marginal likelihood of one group within the hard clustering.  There will be $M = k_1\times\cdots\times k_p$ such groups, indexed by $m$.  The log marginal likelihood for the $m^{th}$ group is then:
\begin{align*}
& \frac{N_m}{2} \: log(\pi) - \frac{1}{2}log(N_m + 1) \\
     &+ log \: \Gamma\Big(\frac{N_m + \delta_t}{2}\Big) 
     - log \:\Gamma\Big(\frac{\delta_t}{2}\Big) \\
     &+ \frac{\delta_t}{2} \: log(\gamma_t) \\
     &- \frac{1}{2}(N_m + \delta_t) \: log(Y_m^T Y_m - \frac{(Y_m^T J_{N_m})^2}{N_m+1} + \gamma_t),
\end{align*}
where $Y_m$ is the vector of responses and $N_m$ is the number of observations in group $m$.
The product of these $M$ approximated marginal likelihoods drives a stochastic search through the space of clusterings.  
Moves are either ``split'' moves that increase $k_j$ or ``merge'' moves that decrease $k_j$.  At the conclusion of this search, the inclusion probabilities, or the proportion of time that the separate $k_j$ are greater than 1 in the course of the search, give an indication of the importance of the corresponding feature to the conditional distribution.  This approach is similar to that presented in \citet{george_mcculloch_1997}.

In the first stage, we examine each of the $p$ features in isolation.  Since it is then feasible (for $d_j \le 5$) to encapsulate the entire stochastic search of corresponding split and merge moves in a discrete time Markov chain, this step proceeds very quickly.  This can be done in an embarrassingly parallel fashion, but experimentation at $p=5000$ where $d_j=4$ for all $j$ showed that the computation of each inclusion probability required 0.3s and so serial computation was not overly burdensome.  We did investigate a marginal likelihood computation that made fewer simplifying assumptions and relied on numerical approximations.  This approach did not produce notably different results and gave a tenfold increase in computational time.

We use the inclusion probabilities from this single-site pass to define a permutation of the features based upon decreasing order of these inclusion probabilities.  We also impose a cutoff from the first stage, including only those features with inclusion probability greater than some value, typically $0.5$.  The cutoff may also be determined by a limit on the size of the space we wish to consider.  The re-ordering before the second stage of variable selection combats the tendency of the stochastic search to jump from simple clusterings to complex clusterings with similar or slightly degraded marginal likelihoods.  If the best candidates from the first-pass search are considered before weaker candidates, the second-pass search performs better.  Depending on the model under consideration, we can perform stochastic search over blocks of features at a time rather than the entire set.
The second stage of variable selection uses a sequential stochastic search variable selection, proceeding for a moderate number of iterations to produce a second set of inclusion probabilities.  This uses the same approximated marginal likelihood approach as in the individual feature assessment.  Features with inclusion probabilities exceeding the cutoff value of $0.5$ are then used in the Gibbs sampling step.

The Gibbs sampler produces a posterior sample according to the steps detailed in section \ref{S:FullCond}.  Each element from this MCMC sample defines a model that we can use to produce predicted values and intervals around predicted values for a test set.

\section{SIMULATION STUDY}
To assess the variable selection and prediction performance of the CTF, we conducted a simulation study, varying the number of training observations $N=500,N=1000$ and using a consistent ground truth to produce simulated data sets with total number of features $p=1000$.  In each case, the true model was based on three features at positions 30, 201 and 801, each with $d_j=4$ levels and including three-way interactions among these features.  The combination of feature values is associated with the mean of an underlying Gaussian, and simulated using a common residual variance $\tau$.
\begin{figure}[ht]
\vskip 0.2in
\begin{center}
\centerline{\includegraphics*[width=\columnwidth,viewport=60 60  750 560]{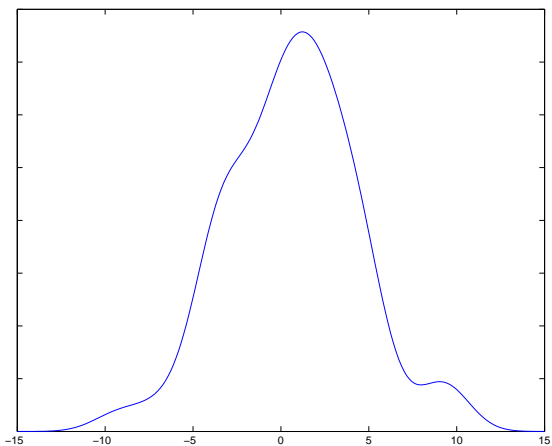}}
\caption{Simulation Study Density.}
\label{F:SimMargDens}
\end{center}
\end{figure} 
The expected number of observations at each combination of levels was equal, giving a population density shown in Figure \ref{F:SimMargDens}.

For each of 20 training sets, we produced selected feature sets and posterior samples based upon the models defined by those sets and the assumed prior structure.  We then used the derived models to make predictions for 20 validation sets drawn from the same underlying true distribution.  As competitor methods we used random forests (RF) and quantile regression random forests (QRF) \citep{meinshausen_2006}; these are implemented in the {\tt randomForest} and {\tt quantregForest} packages in {\tt R}.  BART as implemented in the {\tt BayesTree} package was unable to run to completion on any of the training sets, though we were able to use BART with the real data example in Section \ref{S:RealData}.  

When comparing performance with that of the two competitors, we attempted to
give those competitors what advantages we could.  In the case of RF, this meant that we did two passes over the training data.  The first
pass identified important variables using the {\tt importance} method in the {\tt randomForest} package.  We then fed those variables as a preselected set into a second run.  This generally improved the
MSPE performance of RF.  This option was not available
for QRF, so we could not treat that method in the same manner.
In each of the 20 cases for $p=1000$ and training $N=500$, the CTF outperformed RF on mean square prediction error and showed comparable 95\% coverage proportions to those derived from QRF; this is summarized in Table \ref{T:SimStudyCompare}. The CTF and RF showed comparable accuracy in identifying important features, but RF tended to include many unimportant features.  The table does not show results for metrics that are not appropriate for the particular method.
In contrast, the CTF produced no false positive results.  Both RF and QRF may have suffered due to the strong interactions present in these simulated data.
\begin{table}[t]
\caption{Simulation Study Results.}
\label{T:SimStudyCompare}
\vskip 0.15in
\begin{center}
\begin{small}
\begin{sc}
\begin{tabular}{llllll}
$$          & Training    & $\:$&$\:$&$\:$& $\:$ \\ 
Residual    & Sample & $$ & $$ & $$ & $$ \\
Precision    & Size & Metric & CTF & RF & QRF \\
\hline
\tablespace
1.0       & 500  & MSPE    & 2.27   & 9.76  & -    \\
1.0       & 1000 & MSPE    & 1.40   & 6.78  & -    \\
0.5       & 500  & MSPE    & 3.50   & 10.9  & -    \\
0.5       & 1000 & MSPE    & 2.80   & 7.68  & -    \\
1.0       & 500  & COV    & 0.965   & -     & 0.954    \\
1.0       & 1000 & COV    & 0.958   & -     & 0.959    \\
0.5       & 500  & COV    & 0.953   & -     & 0.950    \\
0.5       & 1000 & COV    & 0.954   & -     & 0.959    \\
\end{tabular}
\end{sc}
\end{small}
\end{center}
\vskip -0.2in
\end{table}
\section{APPLICATION TO REAL DATA}
\label{S:RealData}
To illustrate the utility of this approach, we apply it to a real-world dataset and compare its performance to that of the same competitor methods (RF and QRF).  The dataset concerns DNA damage to instances of different cell lines when exposed to environmental chemicals.  The exposure types are hydrogen peroxide (H2O2) and methyl methane sulfonate (MMS), and the remainder of the feature set is genotype information on 23,210 single nucleotide polymorphisms (SNPs).
\citet{rodriguez_etal_2009} provides extensive details on the original experiments.  100 separate instances of each of 90 cell lines were exposed to each chemical and examined at each of 3 time points (before treatment, immediately after treatment, and a longer time after treatment).  The nature of the measurement is destructive; at the desired time interval, comet assay was performed on each cell and the Olive tail moment \citep{olive_etal_1991} recorded; this assesses the amount of DNA damage in the cell, with higher measurements indicating more damage.   The cells from each line are genetically identical, but the resulting distribution of Olive tail moment (OTM) has a different shape for each cell line.  In addition, these distributions are different at the separate time points; generally, the Olive tail moments are smallest (least damage) before exposure to the chemical, largest (most damage) immediately after exposure, and somewhere in-between after a longer recovery time.  

To develop an appropriate response, we computed empirical quantiles at percentiles $(1/32, 2/32, \ldots, 31/32)$ for each cell line at each of the three time points and then derived a single-number summary $w_{ij}$ to tie these three quantile vectors together for cell line $i$ and exposure $j$.  The summary measure $w_{ij} \in (0,1)$ is the value that minimizes
\begin{align}
\sum_{h=17}^{31}
    \Big |     w_{ij}  Q_{i,N,h} + 
     (1 - w_{ij}) Q_{i,L,h} - 
                  Q_{i,A,h} \Big|
\end{align}
Here, $Q_{i,N,h}$ indicates the $h/32^{th}$ quantile for the $i^{th}$ cell line's Olive tail moment distribution at the \qq{{\bf N}o treatment} time, with corresponding quantities for the \qq{{\bf L}ater} time point and the \qq{immediately {\bf A}fter} time point.  The use of only the higher quantiles reflects our desire to learn more about the extremes of DNA repair.  We used a logit transform to derive our final response $y_{ij} = log(\frac{w_{ij}}{1 - w_{ij}})$; this is appropriate for the assumptions of the model.  Negative values of the response indicate that the OTM distribution long after treatment is closer to the distribution right after treatment; positive values indicate that the \qq{long after} distribution is closer to the distribution before treatment.  

The researchers genotyped the cell lines at 49,428 individual SNPs, each of which had previously been associated with some aspect of DNA repair.  Given the small number of cell lines and the fact that many individuals have two copies of the major allele for these SNPs, many of the SNP profiles were identical and many also had no individuals with two copies of the minor allele.  We recoded the genotypes so that 1 indicated at most one copy of the major allele and 2 indicated two copies of the major allele.  After recoding, we reduced the feature set to those SNPs with distinct profiles, leaving 23,210 SNPs for analysis.

We used leave-one-out cross-validation to assess the performance of the CTF against that of the three competitors RF, QRF, and BART.  Each model from the CTF is represented by an MCMC chain, so for each iterate we developed expected values and 95\% prediction intervals for the left-out observation.  We ran the variable selection chain for 5,000 burn-in iterations and computed inclusion probabilities from 10,000 samples.  We ran the MCMC chain for 40,000 burn-in iterations and retained a sample of 20,000 iterations.  Autocorrelation diagnostics indicated an effective sample size of 15,000.  We used the same burn-in and posterior sample sizes for BART.  As in the simulation study, we used the results from a first run of RF to seed a final run of RF.
The CTF showed consistent selection of the treatment (H2O2 or MMS) as the most
important feature and selected a set of four SNPs (IGFBP5, TGFBR3, CHC1L, XPA)
as features; information about these SNPS is summarized in Table \ref{T:SNPS}.  In contrast, RF chose the treatment variable in 56 of the 180 cross-validation scenarios and did not consistently identify any other features.  The CTF has a higher computational time requirement and took approximately twenty times as long as RF or QRF to estimate a model.  Nevertheless, the improved performance is attractive.
\begin{table}[t]
\caption{Details of SNPs Included In the Model.}
\label{T:SNPS}
\begin{center}
\begin{small}
\begin{sc}
\begin{tabular}{llll}
$\:$   & $\:$& Chromosome \\
Gene   & SNP & position \\ 
\hline
\tablespace
IGFBP5 & rs11575170 & 217256085 \\
TGFBR3 & rs17880594 & 92118885  \\
CHC1L  & rs9331997 & 47986441   \\
XPA    & rs3176745 & 99478631
\end{tabular}
\end{sc}
\end{small}
\end{center}
\end{table}

Comparison with the competitor methods showed patterns similar to the simulation study; Table \ref{T:JTDCompare} compares the results from each method.  The interactions between the treatment and the various SNPs may be weak enough that they do not contribute to the same elevated MSPE that RF demonstrated in the simulation study.  Even though the MSPE for RF was close to that for the CTF, the CTF was able to achieve lower MSPE while not sacrificing coverage performance.
\begin{table}[t]
\caption{Toxicology Data Results.}
\label{T:JTDCompare}
\begin{center}
\begin{small}
\begin{sc}
\begin{tabular}{rllll}
Metric        & CTF     & RF      & QRF   & BART  \\ 
\hline               
\tablespace
MSPE          & 0.263   & 0.353   & -     & 0.425 \\
95\% Coverage & 0.961   & -       & 0.928 & 0.817 \\
\end{tabular}
\end{sc}
\end{small}
\end{center}
\vskip -0.1in
\end{table}
\begin{figure}[ht]
\begin{center}
\centerline{\includegraphics*[width=\columnwidth,viewport=20 20  400 420]{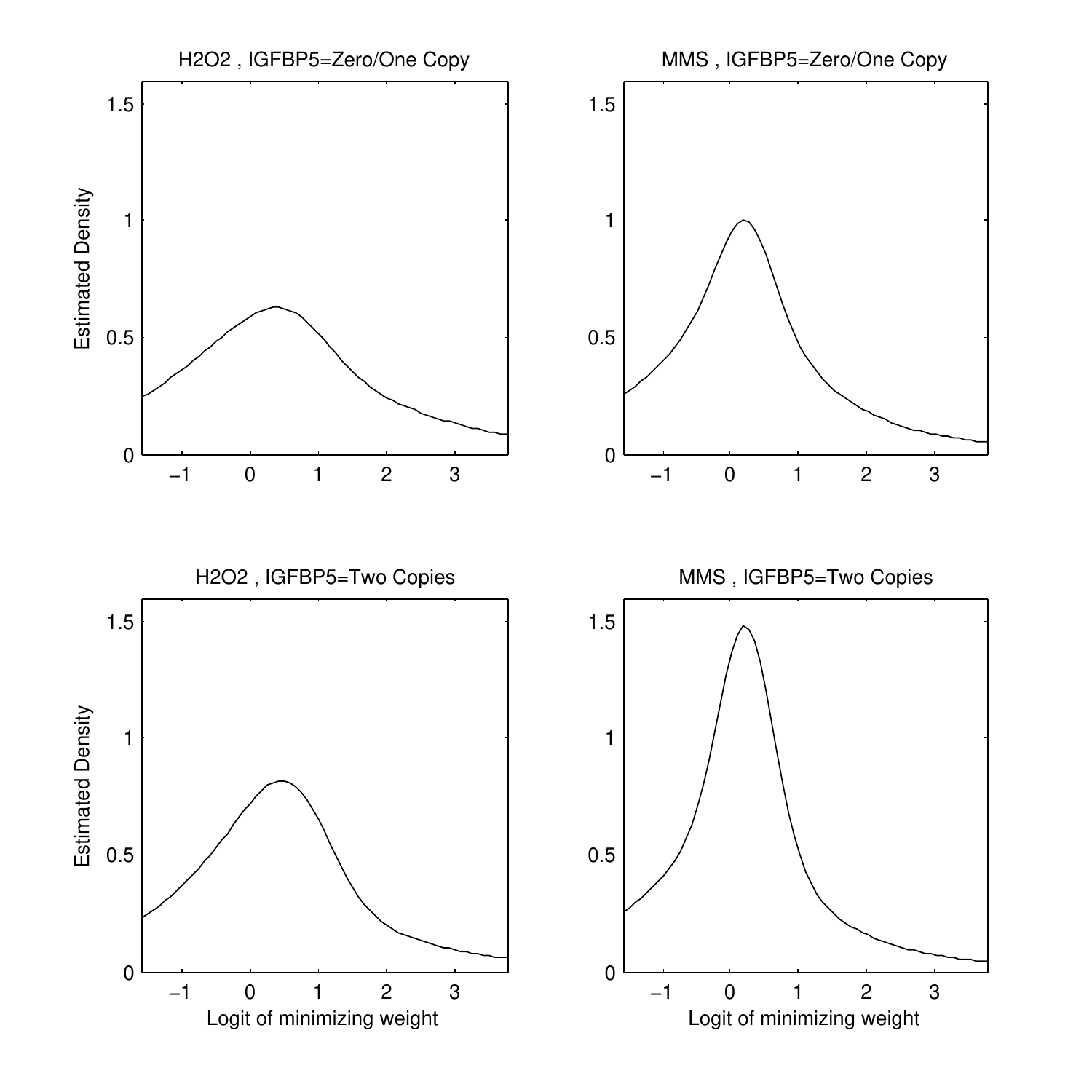}}
\vskip 0.1in
\caption{Selected Conditional Densities For Estimated Model of Toxicology Data.}
\label{F:JTDCOND}
\end{center}
\end{figure} 
Figure \ref{F:JTDCOND} shows estimated conditional densities for selected levels of the treatment and of the IGFBP5 SNP, illustrating how the conditional density changes in more than the conditional mean when the feature vector changes.  In this case, the interaction between MMS treatment and two copies of the major allele for this IGFBP5 SNP tightens the density noticeably, while it has a more muted impact on the conditional mean.  The change is less dramatic under the exposure to H2O2.  In this setting, the shift in the mean response as treatment and genetic profile change is less interesting than the difference in conditional variance; under treatment with H2O2, the mean response is slightly different than under treatment with MMS, but the tail probabilities are noticeably different.
\section{DISCUSSION}
\label{S:Discussion}
We have presented a novel method for flexible conditional density regression in the common case of a continuous response and categorical features.  
The simulation study and real data example suggest that this conditional tensor factorization method can have better performance than general modeling tools when there is substantial interaction between the features of interest.
The CTF does have a higher computational time requirement than the competitor methods, but
the improvement in prediction accuracy and coverage still make the CTF an attractive method.
In addition, a distinct advantage of the CTF is its ability to produce conditional density estimates.
This property of the CTF provides insight beyond a simple conditional expectation and makes it possible to answer more complex questions about the relationship between the response and the features.

\subsubsection*{References}
\begingroup
\renewcommand{\section}[2]{}%
\bibliography{fixctf}
\bibliographystyle{icml2013}
\endgroup 

\end{document}